\documentclass{article}

\usepackage{microtype}
\usepackage{graphicx}
\usepackage{subfigure}
\usepackage{booktabs} 

\usepackage{hyperref}

\usepackage[accepted]{icml2023}

\usepackage{amsmath}
\usepackage{amssymb}
\usepackage{mathtools}
\usepackage{amsthm}
\DeclareMathOperator*{\argmin}{argmin}

\usepackage[capitalize,noabbrev]{cleveref}

\theoremstyle{plain}
\newtheorem{theorem}{Theorem}[section]

\theoremstyle{definition}
\newtheorem{definition}[theorem]{Definition}

\theoremstyle{remark}

\usepackage[textsize=tiny]{todonotes}

\icmltitlerunning{Understanding Predictive Coding as an Adaptive Trust-Region Method}

\begin{document}

\twocolumn[
\icmltitle{Understanding Predictive Coding as an Adaptive Trust-Region Method}



\icmlsetsymbol{equal}{*}

\begin{icmlauthorlist}
\icmlauthor{Francesco Innocenti}{sch}
\icmlauthor{Ryan Singh}{sch}
\icmlauthor{Christopher L. Buckley}{sch}
\end{icmlauthorlist}

\icmlaffiliation{sch}{School of Engineering and Informatics, University of Sussex, Brighton, UK}

\icmlcorrespondingauthor{Francesco Innocenti}{F.Innocenti@sussex.ac.uk}

\icmlkeywords{Predictive Coding, Backpropagation, Trust Region, Saddles, Fisher Information, Local Learning, Inference Learning}

\vskip 0.3in
]



\printAffiliationsAndNotice{} 

\begin{abstract}
Predictive coding (PC) is a brain-inspired local learning algorithm that has recently been suggested to provide advantages over backpropagation (BP) in biologically relevant scenarios. While theoretical work has mainly focused on showing how PC can approximate BP in various limits, the putative benefits of ``natural'' PC are less understood. Here we develop a theory of PC as an \textit{adaptive trust-region} (TR) algorithm that uses second-order information. We show that the learning dynamics of PC can be interpreted as interpolating between BP's loss gradient direction and a TR direction found by the PC inference dynamics. Our theory suggests that PC should escape saddle points faster than BP, a prediction which we prove in a shallow linear model and support with experiments on deeper networks. This work lays a foundation for understanding PC in deep and wide networks.
\end{abstract}

\section{Introduction}
\label{intro}

Predictive coding (PC) is a long-standing theory of cortical function belonging to the wider class of  bio-inspired local learning algorithms (e.g., \citealt{lillicrap2016random, scellier2017equilibrium, ororbia2019biologically, meulemans2020theoretical, payeur2021burst, dellaferrera2022error, hinton2022forward})  which has recently been gaining traction as a biologically plausible alternative to backpropagation \cite{millidge2021predictive, millidge2022predictivereview}. The basic premise of PC is that the brain is constantly trying to minimise prediction errors.

In recent years, there has been considerable effort attempting to relate PC to BP. This work started with \citet{whittington2017approximation}, who showed that PC can approximate the gradients computed by BP on multi-layer perceptrons (MLPs) when the influence of the prior is upweighted relative to the observations. \citet{millidge2022predictive} generalised this result to arbitrary computational graphs including convolutional and recurrent neural networks. A variation of PC, in which weights are updated at precisely timed inference steps, was later shown to be equivalent to BP on MLPs \cite{song2020can}, a result further generalised by \citet{salvatori2021predictive} and \citet{rosenbaum2022relationship}. Finally, \citet{millidge2022backpropagation} unified these and other approximation results as certain equilibrium properties of energy-based models (EBMs).

On the other hand, the ways in which PC, in its natural regime, differ from BP are much less understood. \citet{song2022inferring} proposed that PC, and EBMs more generally, implement a fundamentally different principle of credit assignment called ``prospective configuration''. According to this principle, neurons first change their activity to better predict the output target and then update their weights to consolidate that activity pattern. Based on empirical results, \citet{song2022inferring} suggested that PC can outperform BP in biologically realistic tasks including online and continual learning.

Recent work has started to provide a theoretical basis for this principle. For example, \citet{millidge2022theoretical} showed that (i) in the linear case the PC inference equilibrium can be interpreted as an average of BP's feedforward pass values and the local targets computed by target propagation (TP; \citealt{meulemans2020theoretical}), and that (ii) any critical point of the PC energy function is also a critical point of the BP loss. In the online (mini-batches of size one) case, \citet{alonso2022theoretical} proved that PC approximates implicit gradient descent under specific rescalings of the layer activations. Nevertheless, the fundamental relationship between PC and BP remains to be fully elucidated.

Here, we show that PC can be usefully understood as a form of \textit{adaptive trust-region} (TR) method that exploits second-order information. In particular, we show that the inference stage of PC can be thought of as solving a TR problem on the BP loss using an adaptive, second-order trust region defined by the Fisher information of the generative model. The learning dynamics of PC can then be interpreted as interpolating between the direction of BP's loss gradient and the direction of the TR inference solution, depending on the strength of Fisher information. Our theory suggests that PC should escape saddles faster than BP, a well-known property of TR methods \cite{conn2000trust, dauphin2014identifying, yuan2015recent, levy2016power, murray2019revisiting}. We confirm this prediction in a toy model (Section \ref{toy}) and provide further supporting evidence on deeper networks (\cref{experiments}).

The rest of the paper is structured as follows. After some relevant background on PC and trust region methods (Section \ref{prelim}), we build intuition for the differences between PC and BP by studying a toy network (Section \ref{toy}). We then present our theoretical analysis of PC as a trust-region method (\cref{pc_trust}), followed by some experiments consistent with our theory (\cref{experiments}). We conclude with a brief discussion of our results as well as directions for future research (\cref{discussion}). Notation, theorems, derivations and experiment details are all presented in \cref{appendix}.

\section{Preliminaries} \label{prelim}

\subsection{Predictive coding networks (PCNs)} \label{pcns}
PCNs are energy-based models that implement a hierarchical Gaussian generative model of the data \cite{millidge2021predictive}. Each layer (and neuron within a layer) of a PCN can change its activity $\{ \mathbf{z}_\ell \}_{\ell=0}^L$ and weights $\{ W_\ell \}_{\ell=1}^L$ to minimise its local prediction errors. More formally, PCNs minimise an energy function, called the free energy, that can be reduced to a sum of squares across layers
\begin{equation}
    \mathcal{F} = \sum_{\ell=1}^L \frac{1}{2} \Pi_\ell \big( \mathbf{z}_\ell - f_\ell(W_\ell \mathbf{z}_{\ell-1}) \big)^2
    \label{eq1}
\end{equation}
where $\Pi_\ell$ are layer-wise precision (or inverse covariance) matrices and $f_\ell$ is some activation function.

Here we consider PCNs trained in a so-called ``discriminative'' direction, which have been shown to recapitulate the performance of BP on small-to-medium machine learning tasks \cite{millidge2022predictivereview} and suggested to provide additional benefits in more biologically realistic settings \cite{song2022inferring}, although at a much higher computational inference cost. To train a discriminative PCN, the bottom layer is clamped to data (e.g., labels), $\mathbf{z}_L = \mathbf{x}$, while the top layer is fixed to some ``prior'' (e.g., images), $\mathbf{z}_0 = \mathbf{y}$. The energy (Eq. \ref{eq1}) is minimised in two phases, first w.r.t. the neural activities (inference) and then w.r.t. the weights (learning)
\begin{align}
    &\textit{Inference:} \quad \Delta \mathbf{z}_\ell =  - \eta \frac{\partial \mathcal{F}}{\partial \mathbf{z}_\ell}
    \label{eq2} \\
    &\textit{Learning:} \quad \Delta W_\ell = - \alpha \frac{\partial \mathcal{F}}{\partial W_\ell}
    \label{eq3}
\end{align}
where $\eta, \alpha$ are the respective step sizes. In practice, at every step of training, this minimisation is performed by running the inference dynamics to equilibrium $\Delta \mathbf{z}_\ell \approx 0$, followed by a single weight (e.g., GD) update $\frac{\partial \mathcal{F}}{\partial W_\ell}|_{\mathbf{z}^*}$.

\subsection{Trust region (TR) methods} \label{tr}
A general TR problem \cite{conn2000trust, dauphin2014identifying, yuan2015recent} can be formulated as
\begin{equation}
    \argmin_{\mathbf{z}} f(\mathbf{z}) \quad \text{s.t.} \quad \Delta \mathbf{z} \in \Delta\mathbf{z}^TA\Delta\mathbf{z}
    \label{eq4}
\end{equation}
where $\Delta\mathbf{z}^TA\Delta\mathbf{z}$ defines the geometry and size of the TR, with $A$ being some positive-definite matrix. Special TR algorithms can be derived by (i) different (Taylor) approximations of $f(\mathbf{z})$, (ii) different geometries induced by $A$, and by (iii) whether $A$ depends on the current state of the parameters and is therefore in some sense adaptive.

\section{A Toy Model} \label{toy}

\begin{figure}[h]
    \vskip 0.2in
    \begin{center}
        \centerline{\includegraphics[width=\columnwidth]{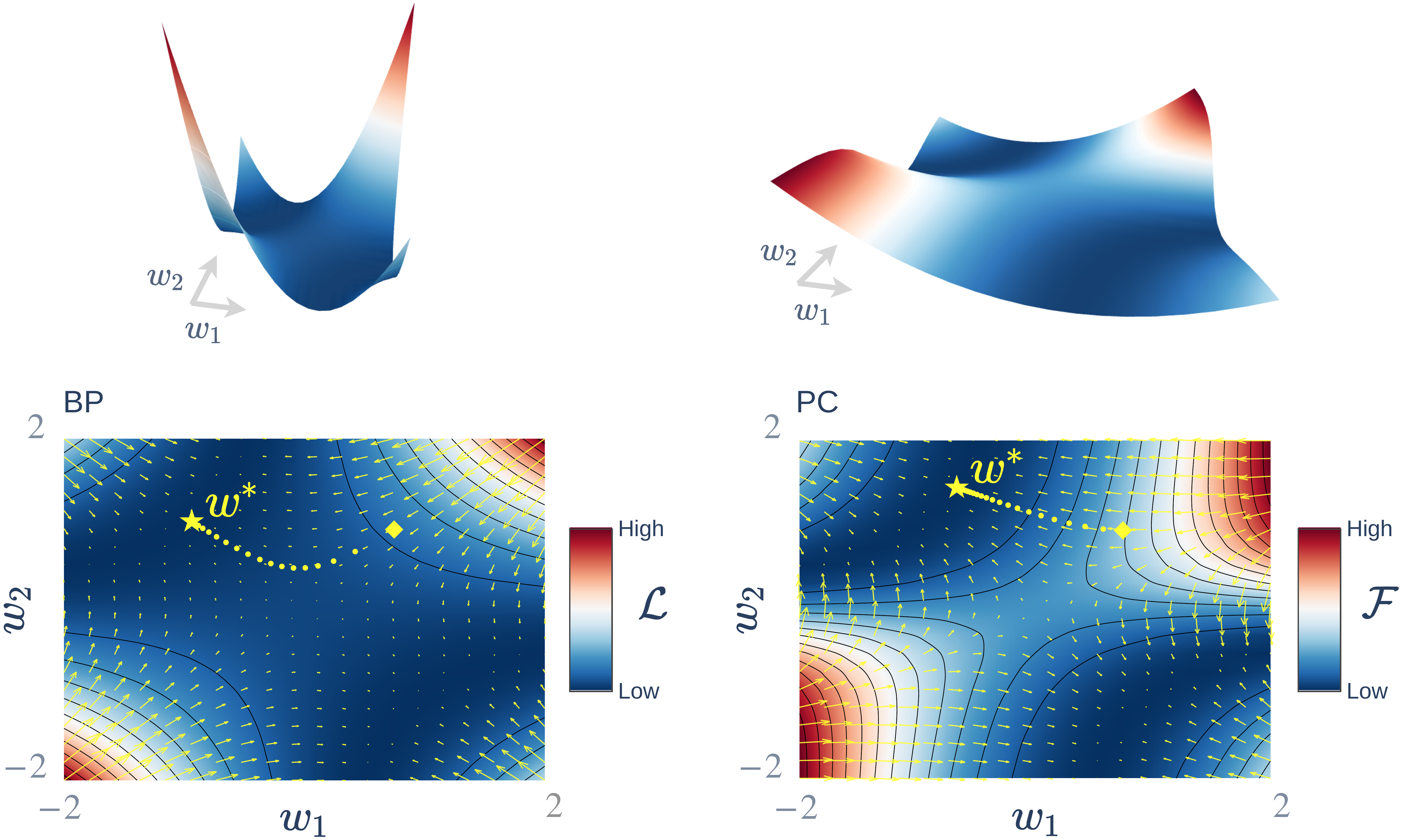}}
        \caption{\textbf{Landscape geometry and GD dynamics of BP vs PC on example 1MLP.} Training loss and energy landscapes of an example 1MLP trained with BP (\textit{left}) and PC (\textit{right}), plotted both as surfaces (\textit{top}) and contours with superimposed gradient fields (\textit{bottom}). Surfaces are plotted at the same scale for comparison, and vector fields are standardised for visualisation. The energy landscape of PC is plotted at the inference equilibrium $\mathcal{F}|_{z=z^*}$ (see \cref{toy_exp} for more details).}
        \label{fig1}
    \end{center}
    \vskip -0.2in
\end{figure}
Here we study a toy model, an MLP with a single linear hidden unit (1MLP) $f(x) = w_2w_1x$, which allows us to compare BP and PC exactly. An example of the landscape geometries and gradient descent (GD) dynamics of the 1MLP weights trained by BP and PC is shown in \cref{fig1} (see \cref{toy_exp} for details). For BP the landscape is simply the loss landscape, while for PC the landscape is the energy landscape \textit{at the inference equilibrium}.

Even in this minimal setting, we can observe marked qualitative and quantitative differences between the two algorithms. In particular, PC seems to evade the saddle, taking a more direct path to the closest manifold of solutions. This is reflected in the geometry of the equilibrated energy landscape, which shows both a flatter ``trap'' direction leading to the saddle and a more negatively curved ``escape'' direction leading to a valley of solutions. Indeed, we provide proof that PC will escape this saddle faster than BP for any non-degenerate problem (\cref{thm:saddle_escape}).


A second observation is the apparent slowdown of PC near a minimum. In the 1MLP case, we prove that this is because the manifold of minima of the equilibrated energy is \textit{flatter} than that of the loss (\cref{thm:flat_min}). This also means, however, that during training (while the target is clamped) PC will be more robust to weight perturbations near a minimum (see \cref{supp_fig3}), which could be important in more biological, online settings.

Thus, in this toy example, we show that PC inference effectively reshapes the geometry of the weight landscape such that GD (i) escapes the saddle faster and (ii) takes longer to converge close to a minimum while being more robust to perturbations.

\begin{figure}[h]
    \vskip 0.2in
    \begin{center}
        \centerline{\includegraphics[width=0.7\columnwidth]{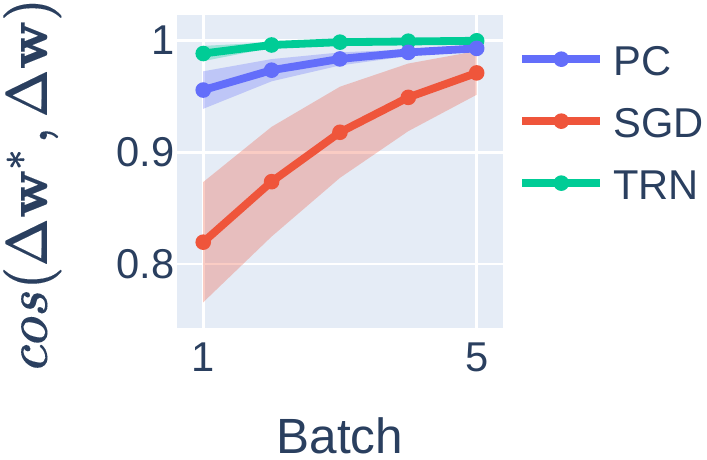}}
        \caption{\textbf{PC weight update direction is closer to optimal than BP on 1MLPs.} For the first $5$ training batches, we plot the mean cosine similarity between the optimal weight direction $\Delta \mathbf{w}^*$ and the update $\Delta \mathbf{w}$ computed by PC (i) $- \nabla_{\mathbf{w}} \mathcal{F}^*$, (ii) BP with SGD, $- \nabla_{\mathbf{w}} \mathcal{L}$, and a (iii) trust-region Newton (TRN) method $- (H + \lambda I)^{-1}\nabla_{\mathbf{w}} \mathcal{L}$ with $\lambda = 2$. Shaded regions indicate the standard error of the mean (SEM) across 10 random weight initialisations.}
        \label{fig2}
    \end{center}
    \vskip -0.2in
\end{figure}

\begin{figure*}[ht]
    \vskip 0.2in
    \begin{center}
        \centerline{\includegraphics[width=0.8\textwidth]{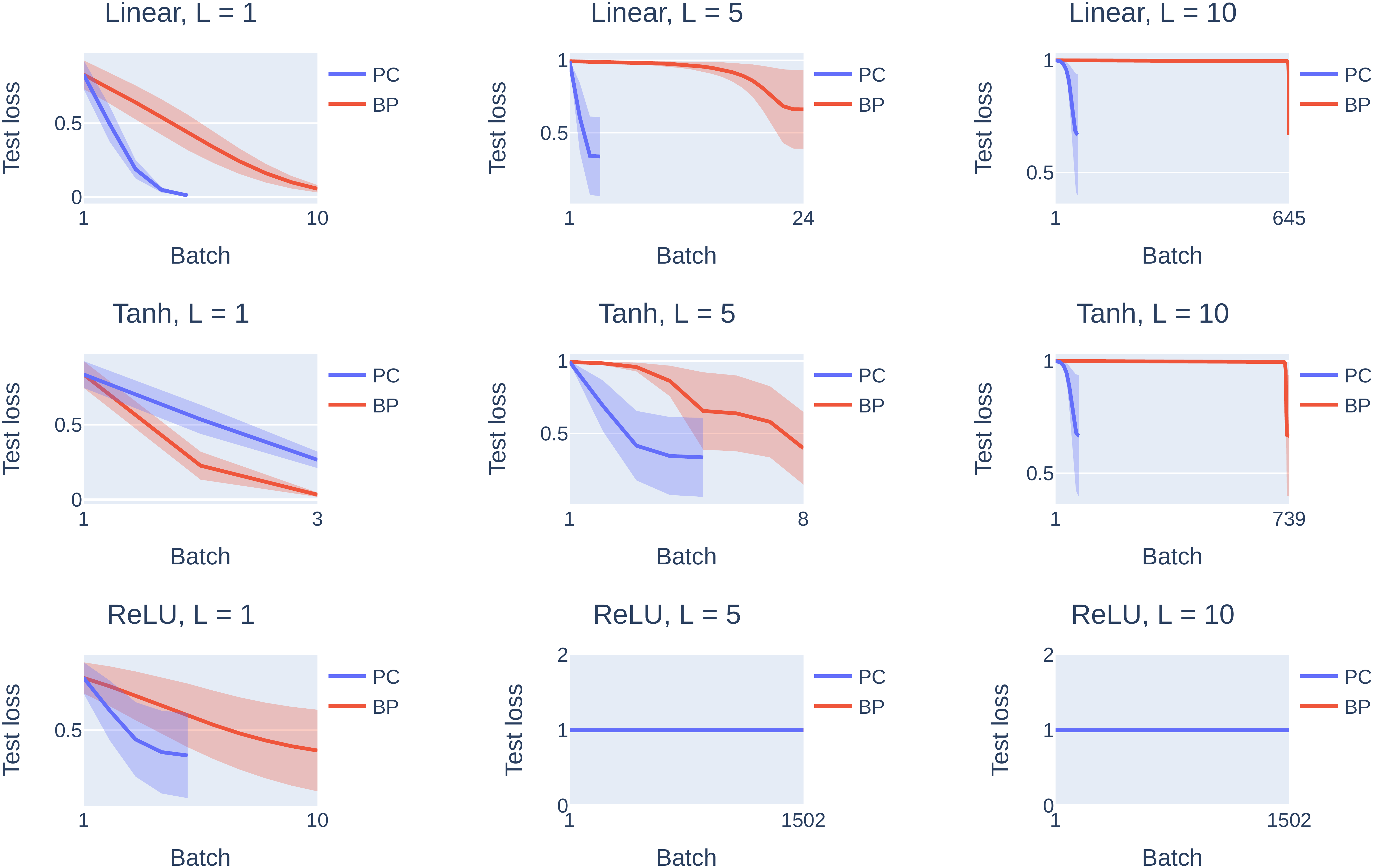}}
        \caption{\textbf{PC can train deeper chains with saddle-inducing activations significantly faster than BP.} Mean test loss of 1D networks (deep chains) trained with BP and PC on the same regression task as in \cref{fig1}. Rows and columns indicate different activation functions (Linear, Tanh and ReLU) and depths $L = \{1, 5, 10\}$, respectively. Each network type was optimised for learning rate, and training was stopped when the loss stopped decreasing (see \cref{dc_exp} for details). Shaded regions represent SEM across 3 different initialisations.}
        \label{fig3}
    \end{center}
    \vskip -0.2in
\end{figure*}

\section{PC as a Trust-Region Method} \label{pc_trust}
Here we show that the inference stage of PC (Eq. \ref{eq2}) solves a TR problem (Eq. \ref{eq4}) on the BP loss with an adaptive, second-order trust region, while the learning stage (Eq. \ref{eq3}) interpolates between GD and the TR solution. In order to make this connection, we perform a second-order Taylor approximation of the free energy (Eq. \ref{eq1}) centred around the feedforward pass values (see \cref{deriv} for a full derivation):
\begin{align}
    \small \mathcal{F}(\mathbf{z}) = \mathcal{L}(\mathbf{z}_t) &+ g_{\mathcal{L}}(\mathbf{z}_t)^T\Delta \mathbf{z} \nonumber \\
    &+ \frac{1}{2} \Delta \mathbf{z}^T \mathcal{I}(\mathbf{z}_t)\Delta \mathbf{z} + \mathcal{O}(\Delta \mathbf{z}^3)
    \label{eq5}
\end{align}
where $g_{\mathcal{L}}$ is the gradient of the loss, $\Delta \mathbf{z} = (\mathbf{z} - \mathbf{z}_t)$, and $\mathcal{I}(\mathbf{z}_t)$ is the Fisher information of the target given by the generative model $p(\mathbf{y} | \mathbf{z})$. This expansion allows us to characterise how the PC energy diverges from the BP loss during inference. We see that Eq. \ref{eq5} defines a TR problem (Eq. \ref{eq4}) with a linear approximation of the loss plus an adaptive, second-order geometry given by $A=\mathcal{I}(\mathbf{z}_t)$. The inference stage of PC (Eq. \ref{eq2}) can thus be thought of as an \textit{adaptive} TR method using second-order information (Eq. \ref{eq5}) with solution 
\begin{equation}
    \mathbf{z}^* \approx \mathbf{z}_t - \mathcal{I}(\mathbf{z}_t)^{-1}g_{\mathcal{L}}(\mathbf{z}_t)
    \label{eq6}
\end{equation}
Because in PC the weights are updated after the activities have converged (\cref{pcns}), when the TR problem is (approximately) solved, we can think of the equilibrated energy landscape as a more ``trustworthy'' landscape than the loss landscape when $\mathcal{I}(\mathbf{z}_t)$ provides useful information. To make this more explicit, we calculate the weight gradient of the energy at the inference solution (see \cref{deriv})
\begin{equation}
    \frac{\partial \mathcal{F}}{\partial W}\biggr|_{\mathbf{z}^*} \approx \frac{\partial \mathbf{z_t}}{\partial W}\mathcal{I}(\mathbf{z}_t)^{-1}g_{\mathcal{L}}(\mathbf{z}_t) + g_{\mathcal{L}}(W)
    \label{eq7}
\end{equation}
where $g_{\mathcal{L}}(W)$ is the loss gradient w.r.t. the weights and $\frac{\partial \mathbf{z_t}}{\partial W}$ is the mapping from activity to weight space. Thus, we see that the gradient on the equilibrated energy interpolates between the direction of the loss gradient and the direction of the TR inference solution mapped back into weight space. 

We can gain insight into these GD dynamics (Eq. \ref{eq7}) by considering the contribution of the Fisher information $\mathcal{I}(\mathbf{z}_t)$. For example, in directions of high Fisher information or model curvature---corresponding to directions of high latent variance---the PC weight update will be biased towards the TR solution. TR methods are known to be better at escaping saddles \cite{conn2000trust, dauphin2014identifying, yuan2015recent, levy2016power, murray2019revisiting}, which is exactly what we observe in the 1MLP case (\cref{toy}). Indeed, we find that the weight direction taken by PC is much closer to that of a TR Newton (TRN) method than BP with GD (see \cref{fig2}). In areas of low Fisher information, on the other hand, PC will tend to look more, but not exactly, like GD, since the curvature will not be zero (unless we are at a critical point where the gradient also vanishes). This is what we seem to observe in the 1MLP case near a minimum, where the model curvature does not seem to provide useful information and slows down convergence. Our theory, then, qualitatively recapitulates the landscape geometry and GD dynamics of PC in the 1MLP case (\cref{toy}).


\section{Experiments} \label{experiments}
Here we report some experiments consistent with the hypothesis, proved for 1MLPs (\cref{thm:saddle_escape}) and suggested by our analysis of PC as a TR method (\cref{pc_trust}), that PC escapes saddles faster than BP. As a first step, we compared the test loss dynamics of BP vs PC on ``deep chains'' $f(x) = w_L f_L(\dots f_1(w_1 x))$. For each algorithm, we selected the best performance across a range of learning rates, to make sure that any differences were not due to inherently different learning rates of PC and BP (see \cref{dc_exp} for more details). Chains with invertible activation functions such as linear or Tanh are invariant to $\prod_{\ell=1}^L 2^{n_{\ell}} = 2^L$ ``sign-flip'', saddle-inducing symmetries \cite{bishop2006pattern, chen1993geometry}, as $n_\ell = 1$ for all layers. Since (S)GD is known to slow down near saddles \cite{dauphin2014identifying, du2017gradient, jin2021nonconvex}, we should expect that BP slows down with depth, while PC should converge more quickly if it indeed avoids saddles faster. On the other hand, we should not expect to find this pattern in chains with non-invertible non-linearities such as ReLU that break sign-flip symmetries (and their associated saddles).

All these predictions were broadly confirmed by our experiments (\cref{fig3}). Overall, we find that PC can train deeper chains with saddle-inducing activations significantly faster than BP. For linear and Tanh activations and chains up to 10 layers, we observe that BP’s convergence with SGD significantly slows down with depth. Indeed, we see the emergence of long plateaus followed by sudden transitions with increased depth, a phenomenon observed in the loss dynamics of deep linear networks \cite{saxe2013exact}. As predicted, we do not see significant speed benefits of PC in ``saddle-breaking'' ReLU networks. We also note that both BP and PC were unable to train very deep chains ($L = 15$), likely due to vanishing/exploding gradients.

\section{Discussion} \label{discussion}
In summary, we showed that PC can be interpreted as an adaptive trust-region method that exploits second-order information. Our theory suggested that PC should escape saddle points faster than BP, a prediction which we verified in 1MLPs and provided further evidence for in deep chains. Preliminary experiments on deep and wide linear networks are also supportive (see \cref{dnn_exp}). In future work, we aim to investigate this saddle behaviour in more depth as well as better understand the nature of the second-order information used by PC to define the trust region geometry.

Our theory can be seen as an important step in providing a more solid theoretical footing to the principle of ``prospective configuration'' \cite{song2020can} and its associated empirical benefits, helping to explain previous results showing faster convergence of PC compared to BP \cite{alonso2022theoretical, song2020can}. We are excited by the possibility of extending this framework to explain, and perhaps uncover, other advantages and disadvantages of PC, such as robustness to small batch sizes and reduced weight interference.

\section*{Code availability}
Code to reproduce all results and plots will be made publicly available at a GitHub URL upon publication of this work. 





\bibliography{example_paper}
\bibliographystyle{icml2023}

\newpage
\appendix
\section{Appendix} \label{appendix}

\subsection{Notation}
Matrices and vectors are denoted with capitals $A$ and small bold characters $\mathbf{v}$, respectively. All vectors are column vectors $[\cdot] \in \mathbb{R}^{n \times 1}$. The gradient and Hessian of any twice differentiable objective function $f : \mathbb{R}^d \rightarrow \mathbb{R}$ w.r.t. $\mathbf{x}$ are denoted as $\nabla_{\mathbf{x}} f$ and $\nabla^2_{\mathbf{x}} f$, respectively. We will often abbreviate them as $g_f(\mathbf{x})$ and $H_f(\mathbf{x})$, omitting the independent variable when it is clear from context. The largest and smallest eigenvalues of the Hessian are $\lambda_{max}(H_f)$ and $\lambda_{min}(H_f)$. We denote any critical point of $f$ where $\nabla_{\mathbf{x}} f = 0$ as $\mathbf{x}^*$.

\subsection{Toy model experiments} \label{toy_exp}
\begin{figure}[h]
    \vskip 0.2in
    \begin{center}
        \centerline{\includegraphics[width=\columnwidth]{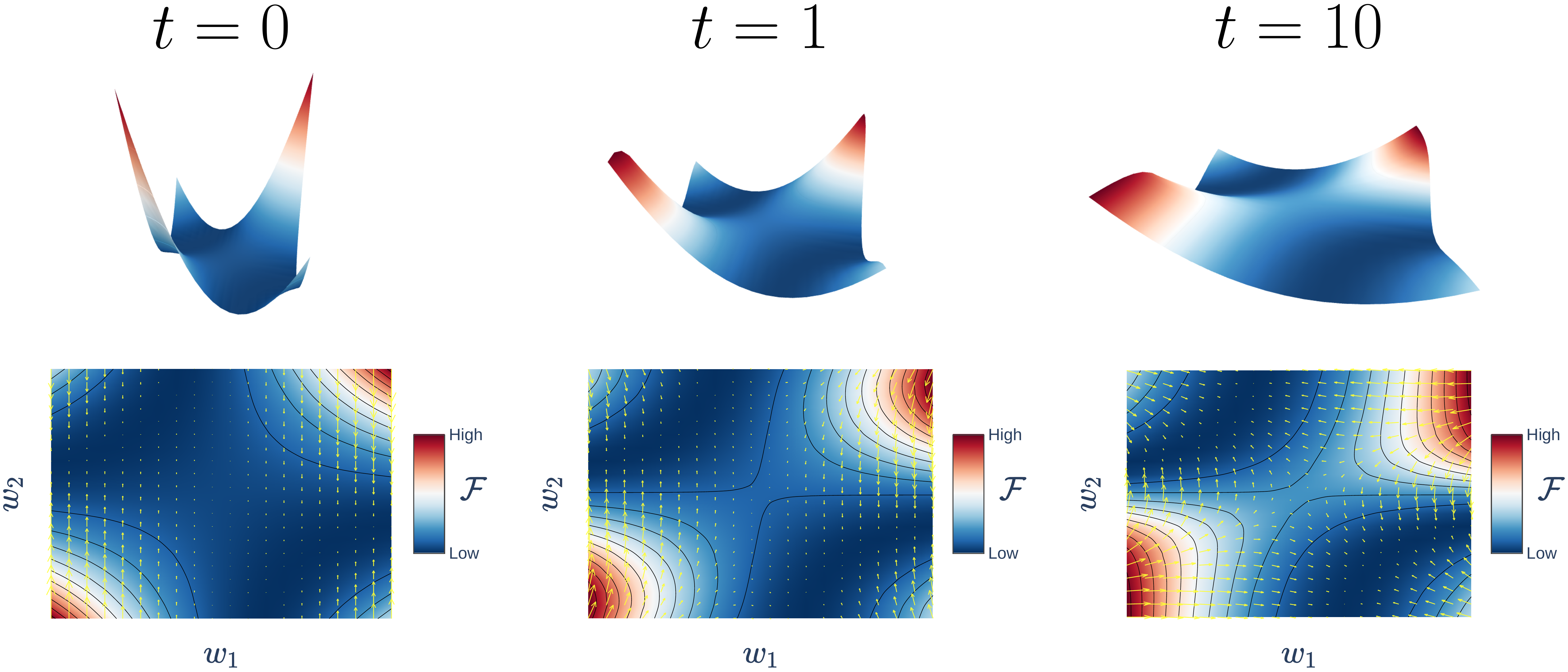}}
        \caption{\textbf{Inference dynamics of PC's energy landscape.} Evolution of the free energy landscape over inference, plotted at initialisation ($t=0$), the first inference step ($t=1$), and equilibrium ($t=10$) for the same 1MLP problem illustrated in \cref{fig1}.}
        \label{supp_fig1}
    \end{center}
    \vskip -0.2in
\end{figure}
1MLPs were trained with BP and PC to predict a simple linear function $y = -x$, where $x \sim \mathcal{N}(1, 0.1)$. We used a uniform weight initialisation $\mathbf{w} \sim \mathcal{U}(-1, 1)$ and SGD with batch size $64$ and learning rate $\alpha = 0.2$ to clearly visualise the algorithms' learning trajectory. Training was stopped when the test loss reached a small tolerance $\mathcal{L}_{test} < 0.001$. For PC, standard Euler integration was used to solve the inference dynamics (Eq. \ref{eq2}), with a feedforward pass initialisation, step size $\eta = 0.1$, and $T = 20$ inference iterations (which were sufficient to reach equilibrium). Precisions were set to one, $p_2 = p_1 = 1$.

To calculate the optimal weight direction at any point during training, we solved for the shortest (Euclidean) distance from the current iterate $\mathbf{w} = (w_1, w_2)$ to the manifold of solutions $\mathbf{w}^* = (w_1^*, \frac{y}{w_1^* x}) = (w_1^*, -\frac{1}{w_1^*})$
\begin{equation}
    D = \sqrt{\left( -\frac{1}{w_1^*} - w_2 \right)^2 + (w_1^* - w_1)^2}
\end{equation}
To minimise this distance, we set the partial derivative of the distance w.r.t. the optimal weight $w_1^*$ to zero
\begin{equation}
    \frac{\partial D}{\partial w_1^*} = \frac{(w_1^*)^4 - (w_1^*)^3w_1 - w_1^* w_2 - 1}{(w_1^*)^3 \sqrt{\left( - \frac{1}{w_1^*} - w_2 \right)^2 + (w_1^* - w_1)^2}} = 0
\end{equation}
Finding the roots of this derivative means solving for the quartic polynomial in the numerator, for which we used numpy. Once solved for the closest solution, we computed the cosine similarity between the optimal direction $\Delta \mathbf{w}^* = (w_1^* - w_1, w_2^* - w_2)$ and the algorithm's GD update at a given point $\Delta \mathbf{w} = - \nabla_{\mathbf{w}} f$
\begin{equation}
    cos(\Delta \mathbf{w}^*, \Delta \mathbf{w}) = \frac{\langle \Delta \mathbf{w}^* \Delta \mathbf{w} \rangle}{\| \Delta \mathbf{w}^* \| \| \Delta \mathbf{w} \|}
\end{equation}
which is simply a normalised dot product.

\subsection{Toy model proofs}
Here we present our two theorems on 1MLPs, showing (i) that PC escapes the saddle point induced by the symmetry in the 1MLP weights faster than BP, and (ii) that the 1MLP mimina of the equilibrated energy are flatter that those of the loss. 

\begin{definition}
\label{def:1mlp_prob}
\textit{1MLP problem.} We define a 1MLP problem as any non-degenerate linear function of the form $y = mx, x, y \neq 0$ that can in principle be learned by a 1MLP $f(x) = w_2w_1x$, where $x, y$ indicate the input and output to the network, respectively.
\end{definition}

\begin{definition}
\label{def:saddle}
\textit{(Strict) saddle.} A critical point $\mathbf{w}^*$ of $f(\mathbf{w})$ where $\nabla f(\mathbf{w}^*) = 0$ is a saddle if the Hessian at that point has at least one positive and one negative eigenvalue, $\lambda_{max}(\nabla^2 f(\mathbf{w}^*)) > 0$, $\lambda_{min}(\nabla^2 f(\mathbf{w}^*)) < 0$. In the literature, these critical points are known as strict or non-degenerate saddles \cite{ge2015escaping, anandkumar2016efficient, jin2017escape}.
\end{definition}

Consider the BP mean squared error loss and PC energy (Eq. \ref{eq1}) associated with a 1MLP problem (\cref{def:1mlp_prob})
\begin{align} 
    \mathcal{L} &= \frac{1}{2} (y - w_2w_1x)^2 \label{eq11} \\
    \mathcal{F} &= \frac{1}{2}(z - w_1x)^2 + \frac{1}{2}(y - w_2z)^2
\end{align} 
where $z$ indicates the value of the hidden unit or latent in PC (which is free to vary). Without loss of generality, we assume a single input-output pair. Note that we can change the sign of the weights without changing the objectives, $f(\mathbf{w}) = f(-\mathbf{w})$. This is known as a ``sign-flip symmetry'' and induces a saddle in the weight landscape \cite{bishop2006pattern, chen1993geometry}. Now recall that we are interested in how PC inference (Eq. \ref{eq2}) affects the learning dynamics (Eq. \ref{eq3}). In the linear case, we can analytically solve for the equilibrium $\frac{\partial \mathcal{F}}{\partial z} = 0, z^* = \frac{w_1x + w_2y}{1 + w_2^2}$ and evaluate the energy at this fixed point
\begin{equation}
    \mathcal{F}^* = \frac{\mathcal{L}}{1 + w_2^2}
    \label{eq13}
\end{equation}
The origin $\mathbf{w}^* = (0, 0)$ is critical point of both the loss and the equilibrated energy since their gradient is zero, $\nabla_{\mathbf{w}} \mathcal{L}(\mathbf{w}^*) = \nabla_{\mathbf{w}} \mathcal{F}^*(\mathbf{w}^*) = \mathbf{0}$. To confirm that this point is a saddle (\cref{def:saddle}), we look at the Hessians
\begin{align}
    H_{\mathcal{L}}(\mathbf{w}^*) &= \begin{bmatrix} 0 & -xy \\ -xy & 0 \end{bmatrix} \\
    H_{\mathcal{F}^*}(\mathbf{w}^*) &= \begin{bmatrix} 0 & -xy \\ -xy & -y^2 \end{bmatrix}
\end{align} 
and see that indeed they both have positive and negative eigenvalues $\lambda(H_{\mathcal{L}}) = \pm xy$, $\lambda(H_{\mathcal{F}^*}) = \frac{1}{2}(-y^2 \pm y\sqrt{4x^2 + y^2})$. Crucially, however, the eigenvalues of the energy are smaller than those of the loss
\begin{equation}
    \begin{cases}
        \lambda_{max}(H_{\mathcal{F}^*}) < \lambda_{max}(H_{\mathcal{L}}) \\
        \lambda_{min}(H_{\mathcal{F}^*}) < \lambda_{min}(H_{\mathcal{L}})
    \end{cases}
    \label{eq16}
\end{equation}
which can be shown by using the fact that the square root of a sum is always smaller than the sum of the square roots, $\sqrt{a^2 + b^2} < \sqrt{a^2} + \sqrt{b^2}$ for $a, b \neq 0$. This result is sufficient to prove that PC will escape the saddle faster than BP, since the near-saddle (S)GD dynamics are controlled by the local curvature. To see this, consider a second-order Taylor expansion of objective $f$ around the saddle
\begin{equation}
    f(\mathbf{w}^* + \Delta \mathbf{w}) \approx f(\mathbf{w}^*) + \frac{1}{2} \Delta \mathbf{w}^T H_f \Delta \mathbf{w}
    \label{eq17}
\end{equation}
where the gradient vanishes. As shown by \citet{lee2016gradient}, taking a gradient descent step of size $\alpha$ from this approximation leads to the following recursive update
\begin{align}
    \begin{split}
        \mathbf{w}_{t+1} &= (I - \alpha H_f)^{t+1} \mathbf{w}_0 \\
        &= \sum_{i=1}^{n_w} (1 - \alpha \lambda_i)^{t+1} \langle e_i \mathbf{w}_0 \rangle e_i
    \end{split}
    \label{eq18}
\end{align} 
where $\mathbf{w}_0 = (\mathbf{w}^* + \Delta \mathbf{w})$, $n_w = 2$ is the number of parameters, and $\{\lambda_i\}_i^{n_w}$ are the Hessian eigenvalues with corresponding eigenvectors $\{e_i\}_i^{n_w}$. We see that GD (or SGD with unbiased noise) will be attracted to, and repelled from, the saddle depending on the degree of curvature along those directions. Because the equilibrated energy has smaller Hessian eigenvalues than the loss at the saddle (Eq. \ref{eq16}), PC will be simultaneously less attracted to and more repelled from it than BP. In other words, the energy saddle turns out to be more ``unstable'', in dynamical systems terms, than the loss saddle.

\begin{theorem}
\label{thm:saddle_escape}
Given any 1MLP problem (\cref{def:1mlp_prob}) which induces a saddle (\cref{def:saddle}) at the origin in weight space, (S)GD on the equilibrated PC energy (Eq. \ref{eq13}) will escape the saddle faster than on the quadratic BP loss (Eq. \ref{eq11}).
\end{theorem}

\begin{figure}[h]
    \vskip 0.2in
    \begin{center}
        \centerline{\includegraphics[width=\columnwidth]{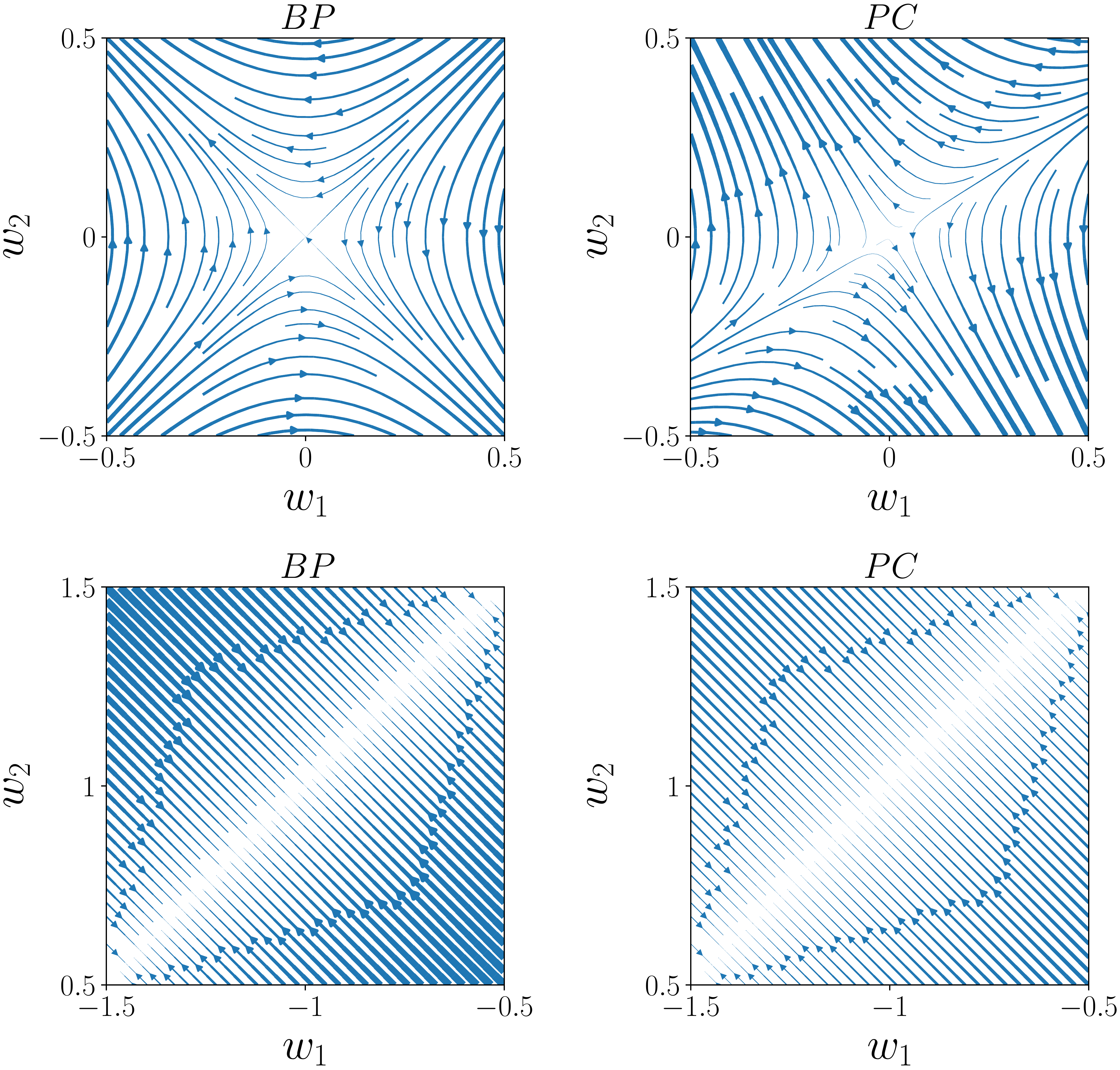}}
        \caption{\textbf{Gradient flow of BP vs PC near different critical points on example 1MLP.} Continuous-time GD dynamics in the vicinity of the saddle (\textit{top}) and an example minimum (\textit{bottom}) of a 1MLP trained with BP and PC on the same regression problem illustrated in \cref{fig1}. Comparing with the discrete and stochastic gradient fields in \cref{fig1}, we note that the continuous dynamics are good approximation.}
        \label{supp_fig2}
    \end{center}
    \vskip -0.2in
\end{figure}

We can also see this by taking the continuous limit of the near-saddle GD dynamics $\alpha \rightarrow 0$ (Eq. \ref{eq18}, \cref{supp_fig2}), leading to the linear ODE system
\begin{equation}
    \dot{\mathbf{w}}(t) = - H_f \mathbf{w}(t)
\end{equation}
with solution $\mathbf{w}(t) = Qe^{\Lambda t}Q^T \mathbf{w}(0)$ and initial condition $\mathbf{w}(0) = (\mathbf{w}^* + \Delta \mathbf{w})$.

Using the same approach, we can also show that any 1MLP global minimum of the equilibriated energy is \textit{flatter} than any corresponding minimum of the loss. Formally, the Hessian eigenvalues of equilibrated energy will also be smaller than those of the loss at any minimum. Because 1MLPs already pose an overparametrised (underdetermined) problem there is no unique solution but rather a manifold. That is, for any value of one weight, there exists only one optimal value of the other, e.g., $\mathbf{w}^* = (\frac{y}{w_2x}, w_2)$. These are also all critical points of both the loss and energy, since their gradient is zero $\nabla_{\mathbf{w}} \mathcal{L}(\mathbf{w}^*) = \nabla_{\mathbf{w}} \mathcal{F}^*(\mathbf{w}^*) = \mathbf{0}$. To verify that this is a manifold of (global) minima, as before we look at the Hessian and see that they both have one zero eigenvalue $\lambda_{min}(H_{\mathcal{L}}) = \lambda_{min}(H_{\mathcal{F}^*}) = 0$ and one positive eigenvalue $\lambda_{max}(H_{\mathcal{L}}) = \frac{w_2^4x^2 + y^2}{w_2^2}$ and $\lambda_{max}(H_{\mathcal{F}^*}) = \frac{w_2^4x^2 + y^2}{w_2^2(1+w_2^2)}$. It is easy to show that the positive curvature of the energy is smaller than that of the loss, $\lambda_{max}(H_{\mathcal{F}^*}) < \lambda_{max}(H_{\mathcal{L}})$.

\begin{theorem}
\label{thm:flat_min}
Given any 1MLP problem (\cref{def:1mlp_prob}), the minima of the equilibrated PC energy are flatter (Eq. \ref{eq13}) than the corresponding minima of the quadratic BP loss (Eq. \ref{eq11}).
\end{theorem}

\begin{figure}[h]
    \vskip 0.2in
    \begin{center}
        \centerline{\includegraphics[width=0.8\columnwidth]{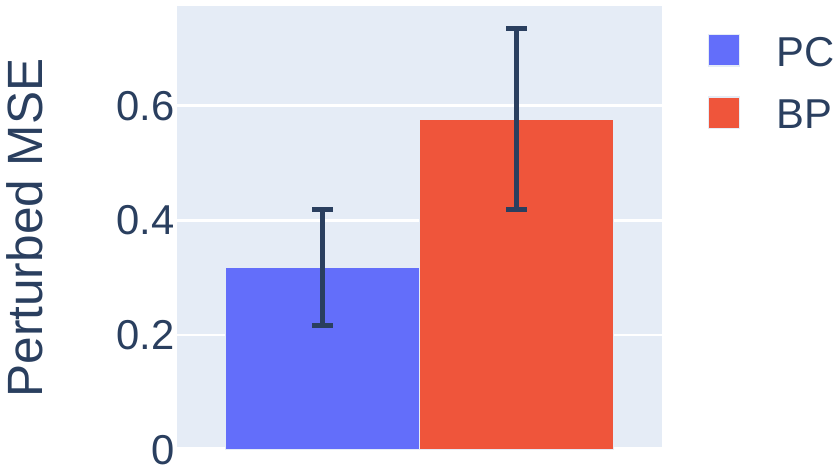}}
        \caption{\textbf{PC is more robust to near-minimum weight perturbations than BP on 1MLP.} Mean squared error (MSE) between output target and weight-perturbed prediction $(y - \hat{y})^2$ of BP and PC trained on the same 1MLP problem illustrated in \cref{fig1}. Weights were perturbed with i.i.d. Gaussian noise $\xi \sim \mathcal{N}(0, 0.5)$. Error bars indicate SEM across 10 different seeds.}
        \label{supp_fig3}
    \end{center}
    \vskip -0.2in
\end{figure}

Performing the same GD analysis as above (Eqs. \ref{eq17}, \ref{eq18}) but around this manifold of minima leads to the conclusion that GD will converge slower than BP when near a minimum, but also be more robust to random weight perturbations where the local approximation holds (\cref{supp_fig3}). As before we can make a similar argument for the continuous case, illustrated in \cref{supp_fig2}.

\subsection{Derivations of theoretical results} \label{deriv}

\textbf{Free energy expansion}.
Recall the free energy is the sum of local prediction errors
\begin{equation}
    \mathcal{F} = \frac{1}{2}(\mathbf{y} - \mathbf{z}_L)^2 + \sum_{\ell=1}^{L-1} \frac{1}{2} \Pi_\ell \big( \mathbf{z}_\ell - f_\ell(W_\ell \mathbf{z}_{\ell-1}) \big)^2
\end{equation}
Let the feedforward activations be defined as $\{ \mathbf{z}_t^{(\ell)} = f^{(\ell)}(\dots f^{(1)}(W^{(1)}\mathbf{z}_t^{(0)})) \}_{\ell=1}^L$ and further we define the difference between $\Delta \mathbf{z} = (\mathbf{z} - \mathbf{z_t})$. Performing a Taylor expansion we see, 
\begin{align}
    \mathcal{F}(\mathbf{z}) =\mathcal{F}(\mathbf{z_t}) &+ J_\mathcal{F}^T(\mathbf{z_t})\Delta \mathbf{z} +  \frac{1}{2}\Delta \mathbf{z}^T H_{\mathcal{F}}(\mathbf{z_t}) \Delta \mathbf{z} \nonumber \\
    &+\mathcal{O}(\Delta \mathbf{z}^3)
\end{align}
and observe the following: $\mathcal{F}(\mathbf{z_t})= \mathcal{L}(\mathbf{z_t})$, $J_\mathcal{F}^T(\mathbf{z_t})=g_{\mathcal{L}}(\mathbf{z_t})$, since in both cases the terms in the sum collapse at the feedforward values. Further, $H_{\mathcal{F}}(\mathbf{z_t}) \approx - \frac{\partial^2 \mathbb{E}_{y, x}\ln p(y, \mathbf{z}, x)}{\partial \mathbf{z}^2 } \mid_{\mathbf{z_t}} = \mathcal{I}(\mathbf{z_t})$ can be seen as the Fisher information of the feedforward values, w.r.t. to the model $p$. Hence:
\begin{align}
    \mathcal{F}(\mathbf{z}) = \mathcal{L}(\mathbf{z_t}) &+ g_\mathcal{L}^T(\mathbf{z_t})\Delta \mathbf{z} +  \frac{1}{2}\Delta \mathbf{z}^T \mathcal{I}(\mathbf{z_t}) \Delta \mathbf{z} \nonumber \\ &+ \mathcal{O}(\Delta \mathbf{z}^3)
\end{align}
\textbf{Approximate inference solution}. 
If we assume $O(\Delta \mathbf{z}^3)$ is a small contribution, we can approximate the inference equilibrium by finding the stationary point of the second-order expansion, yielding
\begin{equation}
    \mathbf{z}^* \approx \mathbf{z_t} - \mathcal{I}(\mathbf{z_t})^{-1}g_{\mathcal{L}}(\mathbf{z_t})
\end{equation}
\textbf{Approximate weight update}.
After the activities converge (at inference equilibrium), PC takes a gradient step on the free energy (\cref{pcns}). In order to find this we first calculate $\frac{\partial \mathcal{F}}{\partial W} = \frac{\partial \mathbf{z_t}}{\partial W}\frac{\partial \mathcal{F}}{\partial \mathbf{z_t}}$:
\begin{equation}
    \frac{\partial \mathcal{F}}{\partial W} = \frac{\partial \mathbf{z_t}}{\partial W}\left[-\Delta \mathbf{z} - \mathcal{I}(\mathbf{z_t})^T\Delta \mathbf{z} + O(\Delta \mathbf{z}^2)\right]
\end{equation}
Finally, plugging in the equilibrium value $\mathbf{z}^*$
\begin{align}
    \frac{\partial \mathcal{F}}{\partial W}\biggr|_{\mathbf{z}^*} &\approx \frac{\partial \mathbf{z_t}}{\partial W}\left[\mathcal{I}(\mathbf{z_t})^{-1}g_{\mathcal{L}}(\mathbf{z_t}) + g_{\mathcal{L}}(\mathbf{z_t})\right] \\
    &\approx \frac{\partial \mathbf{z_t}}{\partial W}\mathcal{I}(\mathbf{z_t})^{-1}g_{\mathcal{L}}(\mathbf{z_t}) + g_{\mathcal{L}}(W)
\end{align}

\subsection{Deep chain experiments} \label{dc_exp}
We trained deep chains on the same regression task for the toy models (\cref{toy_exp}), using SGD with batch size 64. To control for the learning rate $\alpha$, we peformed a grid search over $lrs = \{1e^{-4}, 1e^{-3}, 1e^{-2}, 1e^{-1}, 1e^{-0}\}$ and compared the loss dynamics of the learning rate with the lowest training loss for each algorithm. We recorded training and test loss on every batch from initialisation and stopped training if either (i) the training loss on the current batch was smaller than $\mathcal{L}_{train} < 0.01$, (ii) the average training loss (estimated every 500 batches) did not decrease, or (iii) the loss diverged to infinity (typically because of high learning rates). For PC, we used a similar inference schedule as used by \citet{song2022inferring}, halving the step size $dt = 0.1$ up to two times with maximum $T = 500$ training iterations.

\begin{figure}[H]
    \vskip 0.2in
    \begin{center}
        \centerline{\includegraphics[width=0.6\columnwidth]{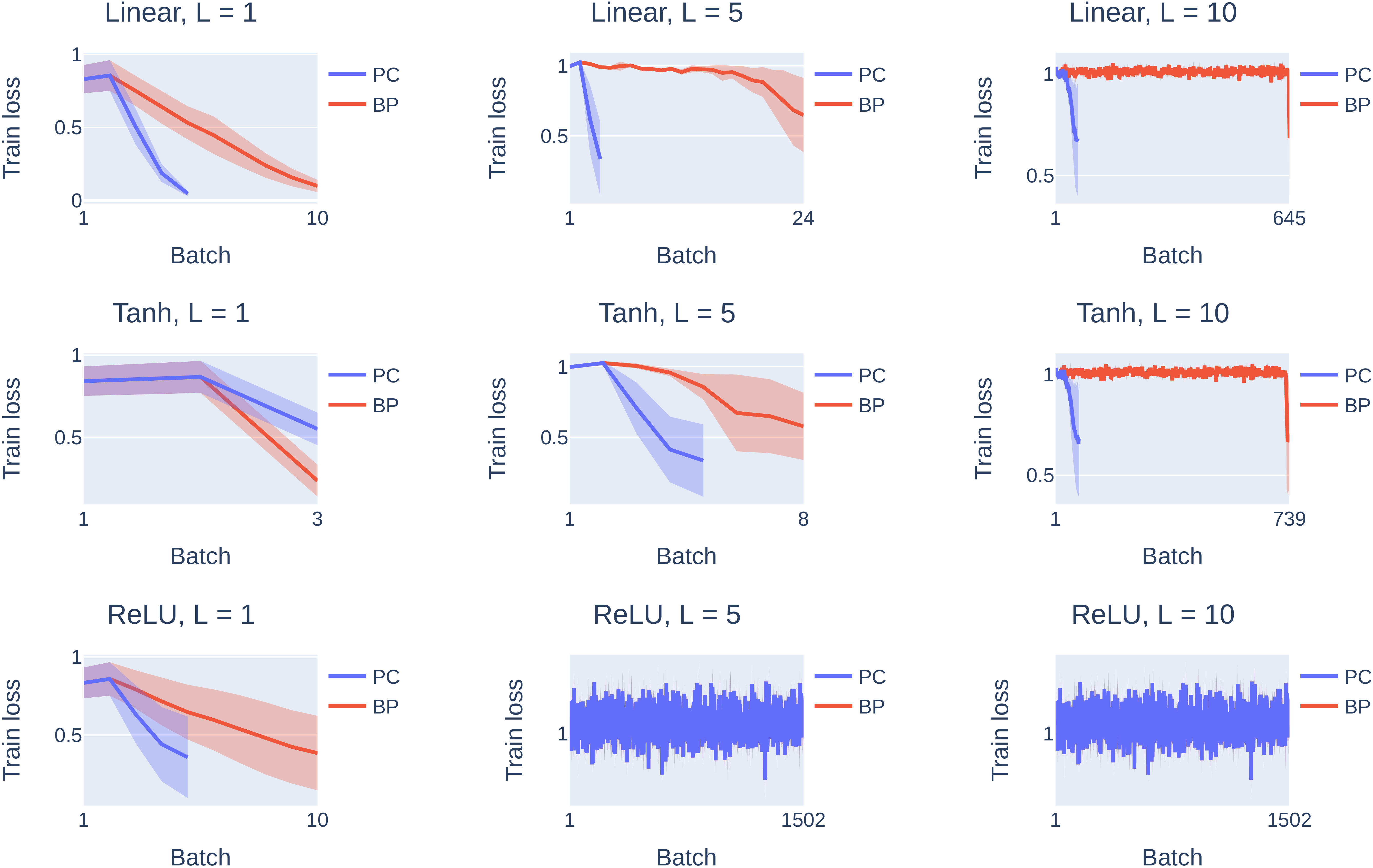}}
        \caption{\textbf{Mean training losses for the deep chain experiments described in \cref{experiments}.}}
        \label{supp_fig4}
    \end{center}
    \vskip -0.2in
\end{figure}

\subsection{Preliminary experiments on DNNs} \label{dnn_exp}
\begin{figure}[h]
    \vskip 0.2in
    \begin{center}
        \centerline{\includegraphics[width=0.9\columnwidth]{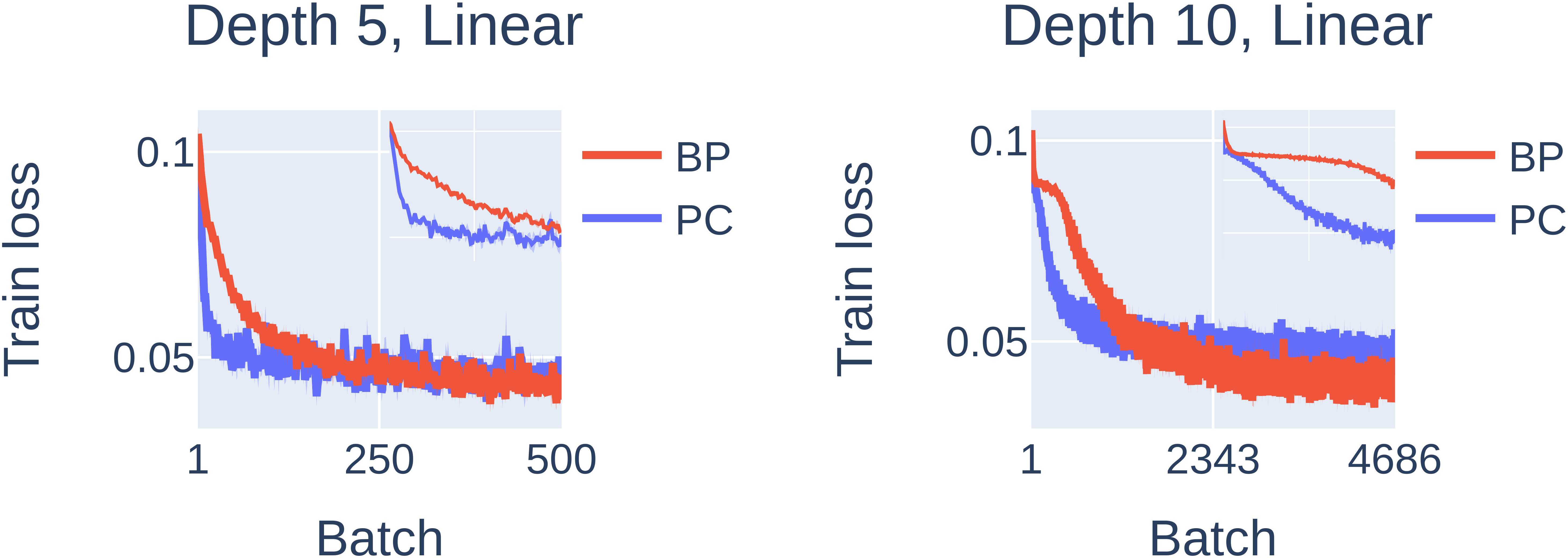}}
        \caption{\textbf{Faster convergence of PC in deep linear networks trained on MNIST.} Mean loss of linear DNNs trained to classify MNIST for 3 random initialisation. Insets highlight PC speed-ups. SEMs are not visible.}
        \label{supp_fig5}
    \end{center}
    \vskip -0.2in
\end{figure}

Linear DNNs of width $n = 500$ were trained on MNIST, using SGD, batch size 64, and the same learning rate grid search for deep chains (\cref{supp_fig5}). As standard, the MNIST images were normalised. Training was stopped if the train loss did not decrease from the previous epoch or diverged to infinity. For PC, all details were the same as deep chains \cref{dc_exp} except for $T = 1000$ maximum inference iterations, used to make sure that failure to train highly overparametrised networks was not due to insufficient inference.

\end{document}